\documentclass[10pt,leqno,fleqn]{article}

\usepackage{graphics}
\usepackage{color}
\usepackage{avm}
\usepackage{tree-dvips}

\avmfont{\scriptsize\sc}
\avmsortfont{\tiny\it}

\definecolor{light}{gray}{.75}

\title{Disambiguating with Controlled Disjunctions}
\author{Philippe Blache\\
2LC - CNRS \\ 1361 route des Lucioles \\ F-06560  Sophia Antipolis \\ {\tt pb@llaor.unice.fr} }
\date{}

\begin{document}
\thispagestyle{empty}
\maketitle
\bibliographystyle{acl}

\begin{abstract}
In this paper, we propose  a disambiguating technique called {\em controlled disjunctions}. This extension of the so-called named disjunctions relies on the relations existing between feature values (covariation, control, etc.). We show that controlled disjunctions can implement different kind of ambiguities in a consistent and homogeneous way. We describe the integration of controlled disjunctions into a HPSG feature structure representation. Finally, we present a direct implementation by means of delayed evaluation and we develop an example within the functionnal programming paradigm.
\end{abstract}

\section{Introduction}

Ambiguity can affect natural language processing at very different levels: it can be very local (e.g. limited to a feature value) or conversely affect entire syntactic structures. But the general disambiguating process remains the same and rely in particular on contextual information. Unfortunately, there exists very few solutions
providing a general account of such a process with a direct and efficient implementation.

In this paper, we propose an approach allowing a general and homogeneous representation of ambiguity and disambiguation relations. This approach constitutes  an extension of {\em named disjunctions} 
(cf. \cite{Dorre90}) and allows a 
direct implementation of relations controlling the disambiguation.

This paper is threefold. In the first part, we approach the question of the representation and we situate our method among the most representative ones. We describe in particular the advantages and drawbacks of named disjunctions and show how different phenomena 
can be described within such a paradigm. In the second part, we propose an analysis of the disambiguating process itself 
and describe some of the control relations existing between the
different parts of the linguistic structure. We integrate the representation of such relations to the named 
disjunction approach: we call this new technique {\em controlled disjunction}.  The last section 
presents the implementation of controlled disjunction which uses different delayed evaluation techniques such as coroutining, constraint propagation 
or residuation.

\section{Representing Ambiguity}

Ambiguity is generally a problem for NLP, but it can also be conceived as a useful device for the representation of particular linguistic information such as homonymy, valency variations, lexical rules, etc. In this perspective, a general representation is very useful.

\subsection{Different needs for representing ambiguity}

Ambiguity can be more or less complex according 
to its scope and there is a general distinction between global and local ambiguities. We can also observe such a difference from a strictly structural point of view: ambiguity can affect the subcomponents of linguistic objects as well as entire structures (or, in a 
feature structure perspective, atomic as well as complex feature values). This is the reason why we distinguish 
two fundamental properties: ({\em i}) the 
interconnection between different subcomponents of an ambiguous structure and  ({\em ii}) the redundancy of 
such structures.

Moreover, we can introduce a certain kind of dynamicity: ambiguity  affects the 
objects differently depending on whether it is related to external values or not. More precisely, certain ambiguities have only 
a minimal effect on the structure to which they belong whereas some others can deeply affect it. A value can be poorly or strongly related 
to the context and the ambiguities are more or less dynamic according to the density of the 
entailed relations.

Figure (\ref{tableau-ambig}) shows several ambiguities involving 
different kind of relations between feature values. In these examples, some ambiguities (e.g. (1.a)) have 
no effect on the other features  
whereas some features are strongly interconnected as in (1.c). Let us remark that the feature type has no consequence on such relations (see for example (1.d) and (1.e)). 

\begin{figure}
\begin{center}
\footnotesize
\begin{tabular}{|l|l|l|l|l|}
\hline
& Unit & Language & Ambiguity	& Relations between values \\
\hline
(1.a) & {\em les} & French &  
	\begin{avm}
		\[\rm \it det $\vee$ \rm \it pro \\
		  \rm \it plur\\
		  \rm \it masc $\vee$ \rm \it fem \]
	\end{avm} & None \\
(1.b) & {\em walks} & English  & 
	\avmoptions{center}\begin{avm}
		\[ \rm \it noun \\
		   \rm \it plur \]
	\end{avm}
	$\vee$
	\avmoptions{center}\begin{avm}
		\[ \rm \it verb \\
		   \rm \it sing \\
		   \rm \it 3rd \]
	\end{avm} & Category and number \\
(1.c) & {\em mobile} & French & 
	\avmoptions{center}\begin{avm}
		\[\rm \it noun \\
		  \rm \it masc \\
		  \rm \it sing \]
	\end{avm}
	$\vee$
	\avmoptions{center}\begin{avm}
		\[\rm \it adj \\
		  \rm \it masc $\vee$ fem \\
		  \rm \it sing \]
	\end{avm} & Category and number \\
(1.d) & {\em die} & German & 
	\begin{avm}
		\[\rm \it det \\
		  \rm \it nom $\vee$ \rm \it acc \\
		  \rm \it plu $\vee$ \avmoptions{center}\begin{avm} \[\rm \it fem \\ \rm \it sing \]\end{avm}\]
	\end{avm} & Gender and number \\
(1.e) & {\em den} & German & 
	\begin{avm}
		\[ \rm \it det \\
		\avmoptions{center}
			\[\rm \it acc \\
			  \rm \it masc \\
			  \rm \it sing \]
		$\vee$
		\avmoptions{center}
			\[\rm \it dat \\
			  \rm \it plu \] \]
	\end{avm} & Case, gender, number \\ 
(1.f) & suffix {\em \_st} & German & 
	\avmoptions{center}\begin{avm}
		\[\rm \it 3rd \\
		  \rm \it sing \]
	\end{avm}
	$\vee$ 	
	\avmoptions{center}\begin{avm}
		\[\rm \it 2nd \\
		  \rm \it plu \]
	\end{avm} & Person and number \\
\hline
\end{tabular}\caption{Examples of interconnection between feature values}
\end{center}
\label{tableau-ambig}
\end{figure}
\setcounter{equation}{1}

The second problem concerns redundancy: when an ambiguity affects major feature values such as category, the result is a set of structures as for (1.b): the ambiguity between the verb and the noun involves in this example two completely 
different linguistic structures. But there are also ambiguities less interconnected with 
other parts of the structure (e.g. (1.d)). In these cases, a common subpart 
of the structure can be factorised. This is particularly useful for the 
representation and the implementation of some descriptive tools such as lexical rules (see \cite{Bredekamp96}). 

An efficient ambiguity representation must take into account the interconnection 
between the different subparts of the structure and allow a factorisation avoiding redundancy. 

\subsection{Different Representations}

The disjunctive representation of ambiguity remains the most natural (see \cite{Karttunen84}, 
\cite{Kay85} or \cite{Kasper90}). However, this approach has several 
drawbacks. 
First of all, if an ambiguity affects more than one atomic value, then a classical disjunction can only represent variation between complete structures.  In other words, such a representation doesn't allow the description of the relations existing 
between the features. In the same way, several approaches (see \cite{Maxwell91} 
\cite{Nakazawa88}, \cite{Ramsay90}, \cite{Dawar90} or \cite{Johnson90}) propose to rewrite 
disjunctions as conjunctions (and negations). This method, in spite of the fact that it can allow 
efficient implementations of some ambiguities, presents the same drawback.

A partial solution, concerning in particular the redundancy problem, can be proposed with the use of {\em named disjunctions} 
(noted hereafter ND; also called {\em distributed disjunctions}). This approach has 
been described by  \cite{Dorre90} and used in several works (see \cite{Krieger93}, 
\cite{Gerdemann95} or \cite{Blache96}). The disjunction here only concerns the variable part of a 
structure (this allows the information factorisation).

A ND binds several disjunctive formulae with an 
index (the name of the disjunction). These formulae are ordered and have the same arity. The variation is controlled by a 
{\em covariancy} mechanism enforcing the simultaneous variation of the ND values: when one disjunct 
in a ND is chosen (i.e. interpreted to true), all the disjuncts occurring at the same rank into the 
other ND formulae also have to be true. Several 
NDs can occur in the same structure. Figure (\ref{ex1}) presents the lexical entry corresponding to 
example (1.d). In the following, the shaded indices represent the names of the disjunctions.

\begin{equation}
\begin{minipage}{7cm}
\begin{avm}
\rm\it den = 
\[spec \[ case \{ \rm\it acc $\vee_{\colorbox{light}{1}}$ dat \} \\
	index \[ gen \{ \rm\it masc $\vee_{\colorbox{light}{1}}$ \_ \} \\
		 num \{\rm\it  sing $\vee_{\colorbox{light}{1}}$ plu \} \] \]\]
\end{avm}
\end{minipage}
\label{ex1}
\end{equation}

This example shows a particular 
case of subvariation. Indeed, the {\em plural dative} determiner stipulates no constraints on the gender. We 
note this subvariation with an {\em anonymous} variable.

However, the ND technique also presents some drawbacks. In particular, covariancy, which allows 
a local representation of ambiguity, is the only way to represent relations between features. Such structures need redundant information as shown in example (\ref{suffix}) (cf. \cite{Krieger93}). Moreover, the semantic of the 
disjunction is lost: this is no longer a representation of a variation between values, but between different set of values. 

\begin{equation}
\begin{minipage}{10cm}
\begin{avm}
\[ morph \[ stemm \\
	    ending \{ \rm \it e $\vee_{\colorbox{light}{1}}$ \rm \it st $\vee_{\colorbox{light}{1}}$ \rm \it t $\vee_{\colorbox{light}{1}}$ \rm \it en \} \] \\
synsem  ...  index \[ per \{ 1 $\vee_{\colorbox{light}{1}}$ 2 $\vee_{\colorbox{light}{1}}$ \{3 $\vee_{\colorbox{light}{2}}$ 2\} $\vee_{\colorbox{light}{1}}$ \{1 $\vee_{\colorbox{light}{1}}$ 3 \} \} \\
num \{ \rm \it sing $\vee_{\colorbox{light}{1}}$ \rm \it sing $\vee_{\colorbox{light}{1}}$ \{\rm \it sing $\vee_{\colorbox{light}{2}}$ \rm \it plu\} $\vee_{\colorbox{light}{1}}$ \rm \it plu \} \] \]
\end{avm}
\end{minipage}
\label{suffix}
\end{equation}

In the same perspective, covariancy forces a symmetrical relation between two feature 
values. In fact, there are relations expressing finer selection relations. This is 
the case in (1.c) in which no covariancy between the part-of-speech and 
gender features occurs: an {\em adjective} is either {\em masculine} or {\em feminine} whereas a {\em noun} is always 
{\em masculine}. A classical ND representation as in example (\ref{suffix}) doesn't account for the fact that the noun 
selects a masculine value for the gender feature, but the reverse is not true. 

\section{Disambiguating}

Generally, 
disjunctions are expanded into a disjunctive normal form and the disambiguation comes to a well-formedness verification. 
In case of incoherence, backtracking is applied and another model is elaborated (i.e. another value is 
choosen). This method can be improved by some techniques (cf. \cite{Maxwell91}), but the basic 
mechanism remains a kind of generate-and-test device: the first feature instantiation leads to the choice 
of an entire structure which has to be validated during the parse.

\subsection{Fundamental Needs}

A natural solution consists in delaying  
the evaluation of the structure consistency. We highlight here two techniques used in our approach: selection constraints (cf. 
\cite{Pulman96}) and coroutining (cf. \cite{vanNoord94}, \cite{Blache96}). 

\cite{Pulman96} represents ambiguities with fixed-arity lists. They are 
controlled by an agreement-like device between two categories, one being controlled by the other. These lists can be interpreted as disjunctions (not 
expanded into a normal form) and their evaluation requires contextual information. 
The problem is that this mechanism can only be represented by a phrase-structure rules, and not at a general level. This is problematic in two 
respects: it is a context-sensitive mechanism and it only works for formalisms using phrase-structure rules (unlike HPSG for example). Moreover, this approach involves the definition of a new kind of 
categorial relations (cf. in Pulman's paper the relation between prepositions and nouns) which have 
no linguistic motivation. Finally, the design of several control relations can rapidly become an exhaustive 
description of all the dependencies.

A coroutining approach applied to named disjunctions is described in \cite{Blache96}. The idea  
is to propose the on-line use of NDs relying on (actual) delayed evaluation. The advantage is a direct 
interpretation of NDs: this representation of ambiguity doesn't use any pre-computation, there are no 
normal form expansion  and the same mechanism is reused for all kind of ambiguities. In this 
approach, NDs are no longer simple ``macros" as described in \cite{Bredekamp96}: 
there is no interpretation of this formalisation into another one and the evaluation is direct. This 
is possible because the factorisation allows the construction of an underspecified structure which can 
be used during the parse. The problem however concerns the disambiguation propagation. More 
precisely, the only control is that of the covariancy which doesn't adress the problem of embedded 
NDs.

An efficient treatment of ambiguity would need a factorised representation allowing the use 
of underspecification together with a fine specification of disambiguating relations. These 
characteristics are fundamental for an implementation avoiding DNF expansion and delaying  
disambiguation as much as possible.

\subsection{Controlled Disjunctions}

The main interest with NDs is the relation between different disjunctive formulae  implementing the disambiguation propagation. This propagation supposes an equivalence relation in the sense that any value belonging 
to a ND can be instantiated and propagates the information to the rest of the ND. But this cannot tackle the 
problem of directed disambiguation relations as shown in  example (\ref{mobile}). In this example, we can say that 
({\em i}) if the word is a noun, then the gender is masculine and ({\em ii}) if the word is feminine, 
then it is an adjective. But the reverse is not true: we cannot deduce anything if the masculine value 
is instantiated. 

\begin{equation}
\begin{minipage}{10cm}
\begin{avm}
\[ cat \[ head \osort{adj}{\node{a}{\[ maj \rm\it Adj \\
				mod \rm \it N \]}}
		$\vee_{\colorbox{light}{1}}$ \hspace{1mm}
		\osort{noun}{\node{c}{\[maj \rm\it Noun \\
				Nform \]}} \\
	valence \{ \<\> $\vee_{\colorbox{light}{1}}$ \[spr \rm\it Det\]\} \] \\
   cont \[ index \[ Gen \{ \{ \rm \it masc $\vee$ \rm \it \node{b}{fem} \} $\vee_{\colorbox{light}{1}}$ \rm \it \node{d}{masc} \} \] \] \]
\end{avm}

\anodecurve[b]{b}[b]{a}{1in}[0.5in]
\anodecurve[b]{c}[t]{d}{1in}[1in]
\end{minipage}
\label{mobile}
\end{equation}
\vspace{6mm}

We propose in the following the introduction of the notion of {\em controlled disjunctions}. They allow an integration of all the control relations relying on the distinction between 
{\em controlling} and {\em controlled} values within the ND framework. 

The first problem being that of covariancy, we identify two kind of 
disjunctions: 
\begin{itemize}
\item {\em Simple disjunctions}: their values can be controlling, controlled or both. 
They don't belong to a set of disjunctive formulae as for classical NDs but are represented in a 
separate ND which has its own name and contains this unique formula. This is the  example case
with the gender value of examples (1.c) or (1.e).

\item {\em Covariant disjunctions}: their values are necessarily both controlling and controlled. 
These formulae are equivalent to classical NDs.
\end{itemize}

Insofar as covariancy is not the only relation between features, the second 
problem is the representation of control relations. In a covariant relation, all values occuring at 
the same rank of a formula are both controlling and controlled. Conversely, the description 
of non-covariant relations (involving simple disjunctions) relies on the distinction between controlling 
and controlled values. In example (1.c), the noun value controls (at least) the masculine, and 
the feminine controls the adjective.

A general solution consists of specifying explicitly all the non-covariant control relations. As said before, {\em controlled disjunctions} (hereafter CD) are a ND extension in the sense that they are named and their formulae are ordered. So, any 
value can be referenced by the name of the disjunction and its rank in the formula. In the case of a 
simple CD, these references specify a unique value whereas for a covariant CD, they specify a set of 
covariant values.

We note this control relation as: $value_{\langle i,j \rangle}$ where $value$ is the name of the controlling value, $i$ is the name of the CD and $j$ is the rank of the controlled value in the formula.

Figure (\ref{ex2}) is a representation in an HPSG notation of the example (1.c).

\begin{equation}
\begin{minipage}{10cm}
\begin{avm}
\rm\it mobile = 
\[cat \[head \colorbox{light}{1}\{ \rm\it noun$_{\langle \colorbox{light}{2},1 \rangle}$, adj \} \\
	valence $\mid$ spr \[ \colorbox{light}{1}\{ \rm\it [Det], [] \} \] \] \\
   index \[ gen \colorbox{light}{2}\{\rm\it masc, fem$_{\langle \colorbox{light}{1},2 \rangle}$ \}  \] \]
\end{avm}
\end{minipage}
\label{ex2}
\end{equation}

This structure contains two controlled disjunctions indexed by \colorbox{light}{\footnotesize 1} and  \colorbox{light}{\footnotesize 2}. Disjunction \colorbox{light}{\footnotesize 1} is {\em covariant} and the instantiation of one of its value entails that of the other 
values of the same rank belonging to the same CD. Disjunction \colorbox{light}{\footnotesize 2} is {\em simple} and specifies two possible values. The control relations are specified by the controlling values. These values can either belong to a covariant or a simple disjunction. In this example, the 
controlling values are  $noun_{\langle \colorbox{light}{\footnotesize 2},1 \rangle}$ and $fem_{\langle \colorbox{light}{\footnotesize 1},2 \rangle}$, 
they control respectively the 1st value of the 2nd disjunction (i.e. $masc$) and the 2nd value of the 
1st disjunction (i.e. $adj$ and all its covariant values).

We can remark that, in the case of typed structures, the elements belonging to a control relation can either be feature 
values or types: this is the case of the first disjunction which concerns the head types $noun$ and 
$adj$. This property can be very useful for the expression of high-level ambiguities.

The example of figure (\ref{CD2}) shows a more complex case with embedded controlled disjunction. But we can observe that the mechanism is the same.

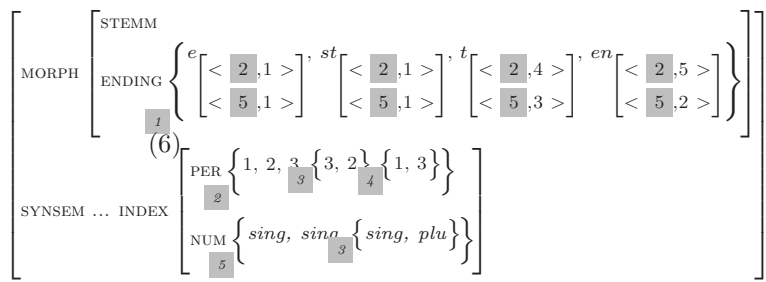
\begin{equation}
\begin{minipage}{14cm}
\newbox\boitea
\setbox\boitea=\hbox{\begin{avm}
\[ < {\colorbox{light}{2}},1 > \\
   < {\colorbox{light}{5}},1 > \]
\end{avm}}

\newbox\boiteb
\setbox\boiteb=\hbox{\begin{avm}
\[ < {\colorbox{light}{2}},1 > \\
   < {\colorbox{light}{5}},1 > \]
\end{avm}}

\newbox\boitec
\setbox\boitec=\hbox{\begin{avm}
\[ < {\colorbox{light}{2}},4 > \\
   < {\colorbox{light}{5}},3 > \]
\end{avm}}

\newbox\boited
\setbox\boited=\hbox{\begin{avm}
\[ < {\colorbox{light}{2}},5 > \\
   < {\colorbox{light}{5}},2 > \]
\end{avm}}

\begin{avm}
\[ morph 
 \[ stemm \\
   ending \osort{\colorbox{light}{1}}{\{ $e_{\box\boitea}$, 
	$st_{\box\boiteb}$, $t_{\box\boitec}$, $en_{\box\boited}$ \}} \] \\
synsem  ...  index \[ per \osort{\colorbox{light}{2}}{\{ 1, 2, 3, \osort{\colorbox{light}{3}}{\{3, 2\}}, \osort{\colorbox{light}{4}}{\{1, 3 \}} \}} \\
num \osort{\colorbox{light}{5}}{\{ \rm \it sing, \rm \it sing, \osort{\colorbox{light}{3}}{\{\rm \it sing, \rm \it plu\}}\}} \] \]
\end{avm}
\end{minipage}
\label{CD2}
\end{equation}

In another way, simple disjunctions, in addition to their controlling capacities, can be 
assimilate to finite domains contraints (in the constraint programming sense). Such constraint are very 
useful for several reasons: they constitute an efficient control on the unification process. Moreover, if no disambiguation can be applied, they propose an approximation of the solution.

\section{Delayed Evaluation}

Delayed evaluation techniques allow us to maintain the disjunctive structure until  disambiguating 
information is found. This has several advantages. First of all, there are no arbitrary choices as with classical 
disjunctive approaches relying on normal forms: all instantiations are guided by contextual 
informations. Moreover, the disambiguation propagation of CD allows us to trigger 
the process at any level: in the case of HPSG, the disambiguation can come from principles, 
schemas or even at the lexical level with structure sharing.

We 
propose in this paper an approach using {\sf LIFE}, a multi-paradigm programming language (cf. 
\cite{Ait-Kaci94}) integrating in particular functionnal facilities to a logic programming framework.

A direct implementation of the CD presents two problems: the different kind 
of relations (covariancy and control) and the control of delaying devices. We propose here a method 
relying on functions which residuate if their arguments are insufficiently 
specified. As soon as these arguments are instantiated, their evaluation is fired. This mechanism is very 
similar to the delayed ones in Prolog ({\tt freeze, block}, etc.) and  correspond in some way to a 
constraint programming approach.

We distinguish for the implementation three kind of relations implemented by three different 
functions. These functions are associated with the values involved in the CD. 
Two functions implements the controlling and controlled relations and one describes the directed (non-covariant) selection.

The representation used for this relies on the following points: 
\begin{itemize}
\item disjunctive formulae are represented with lists,
\item the name of the CD is represented by an integer variable $I$,
\item $I$ also represents the rank of the instantiated disjunct in the list.
\end{itemize}

The residuation depends on ({\em i}) the feature value or ({\em ii}) the rank of the choosen value in 
the list. When a feature value can be disambiguated, its instantiation in the list entails the instantiation 
of its rank. The other values controlled by the same name (i.e. belonging to the same CD) are controlled by a function residuating on the integer variable representing this rank. 
So, when a value is choosen, all the values of the same rank in the disjunction are also instantiated. 
The mechanism is the same with directed selections. In this case, we need a particular function binding two specific values which don't belong to the same CD. These relations are expressed again with 
the integer variable. To summarize ({\em i}) if a value belonging to a CD is instantiated, then its rank 
become known and ({\em ii}) if the rank is known, then the value can be instantiated. Let us describe 
more precisely these functions. In the following, the function {\tt cond(a,b,c)} means that if {\tt a} is 
true, then {\tt b}, otherwise {\tt c}.

\begin{itemize}
\item Function $controlling$ :
\begin{minipage}[t]{7cm}
{\footnotesize
\tt
\begin{tabbing} TeteDe\= Regle\= \kill  
controlling(I,A,L) $\rightarrow$ 	\\
\>	controlling\_followed(1,I,A,L).	\\
controlling\_followed(J,I,A,[X|L]) $\rightarrow$ 	\\
\>	cond(A:==X, A $\mid$ I=J, controlling\_followed(J+1,I,A,L)).
\end{tabbing}}
\end{minipage}

{\tt I} is an integer representing the rank, {\tt A} the feature value to be instantiated and {\tt L} the list 
of possible values. This function residuates on {\tt A}. If {\tt A} is known, its evaluation is fired, the 
function calculates the position of the corresponding value in the list {\tt L} and returns the value 
itself.

\item Function $controlled$ :
\begin{minipage}[t]{7cm}
{\footnotesize
\tt
\begin{tabbing} TeteDe\= Regle\= \kill  
controlled(I,[X|L]) $\rightarrow$ 	\\
\>	cond( \>	I>0, 	\\
\>\>			cond(I=:=1,X,controlled(I-1,L)),	\\
\>\>				false).
\end{tabbing}}
\end{minipage}

The two arguments are the rank (variable {\tt I}) and the list of values ({\tt [X|L]}). The function 
residuates on {\tt I}. When {\tt I} is known, then the function returns the value corresponding to this 
position in the list.

\item Function $selection$ :
\begin{minipage}[t]{7cm}
{\footnotesize
\tt
\begin{tabbing} TeteDe\= Regle\= \kill  
selection (I,X,Y) $\rightarrow$ 	\\
\>	cond( \>	I>0, 	\\
\>\>			cond(I=:=X,Y,true), \\
\>\>			false).
\end{tabbing}}
\end{minipage}

{\tt I} represents the position of a controlling value (and the name of the corresponding CD), {\tt X} 
represents  the value that {\tt I} must have to select the controlled value and {\tt Y} represents the 
rank of the controlled value to be instantiated. This function residuates on {\tt I}. It returns the 
position of the controlled value in case the controlling one has the right value.

\end{itemize}

Let us present now the implementation of the lexical entry of the figure (\ref{ex2}).

\begin{equation}
\begin{minipage}{7cm}

{\footnotesize
\tt
word(mobile,A) :-	\\
\begin{tabular}{lll}
&     A.phon=mobile, \\
&     \node{a}{A.synsem.loc.cat.head=\colorbox{light}{H},} & \node{d}{\it Lists}	\\
&     \node{b}{A.synsem.loc.cat.valence.spr=\colorbox{light}{S},}	\\
&     \node{c}{A.synsem.loc.content.index.gender=\colorbox{light}{G},}	\\
&     \node{e}{H=controlling(I,H,[adj,\colorbox{light}{noun}]),}	\\
&     S=controlling(I,S,[[],[Det]]),	\\
&     \node{f}{G=controlling(J,G,[masc,\colorbox{light}{fem}]),} & \node{g}{\it Control features}	\\
&     H=controlled(I,[adj,noun]),	\\
&     S=controlled(I,[[],[Det]]),	\\
{\it Disjunction}&     G=controlled(J,[masc,fem]),	\\
\node{h}{\it names}&    \node{i}{\colorbox{light}{I}=selection(J,2,1),}	\\
&     \node{j}{\colorbox{light}{J}=selection(I,1,2).}
\end{tabular}}

\nodeconnect[r]{a}[l]{d}
\nodeconnect[r]{b}[l]{d}
\nodeconnect[r]{c}[l]{d}
\nodeconnect[r]{e}[l]{g}
\nodeconnect[r]{f}[l]{g}
\nodeconnect[r]{h}[l]{i}
\nodeconnect[r]{h}[l]{j}
\end{minipage}
\label{life}
\end{equation}

This lexical entry bears two CDs: a covariant one controlled by {\tt I} and a simple one controlled by {\tt J}. 
In this example, all the disjunctive formulae contain values which can be controlled or controlling 
(this is necessarily the case for covariant CD, but not for the simple ones). So the three corresponding 
lists are in the scope of both the $controlling$ and $controlled$ funtions. Moreover, the particular 
directed relations between {\tt noun/masc} and {\tt fem/adj} are implemented by the $selection$ 
function which indicates for example in its first occurrence that if the second value of the CD J is 
instantiated, then the CD {\tt I} must instantiate the first value (i.e. if {\tt J=2}, then {\tt I=1}).

\section{Conclusion}

The {\em controlled disjunction} technique presented in this paper allows a concise representation of disjunctive information, the description of precise variation phenomena (not only covariancy as with ND) together with an efficient implementation relying on underspecification and delayed evaluation. It consitutes an important extension of the named disjunction device: its representation of control phenoma between disjunctive values allow the implementation of all kind of ambiguities.

This approach can be improved in several respects and in particular with consistency tests: verification that disjunctions are exclusive, no cycles in  the dependency net of relations between disjunctive values, etc.   

The implementation of controlled disjunctions is direct using classical delayed evaluation techniques. The example proposed in this paper relies on functional programming, but other logic programming techniques can also be used.

\end{document}